\documentclass[letterpaper, 10 pt, conference]{ieeeconf}
\IEEEoverridecommandlockouts
\usepackage{cite}
\usepackage{amsmath,amssymb,amsfonts,amsthm}
\usepackage{graphicx}
\usepackage{textcomp}
\usepackage{xcolor}
\usepackage{url}
\usepackage{tablefootnote}

\usepackage{algorithm}
\usepackage{algpseudocode}
\usepackage{subcaption}
\usepackage{multirow}
\usepackage{booktabs}
\usepackage{makecell}

\usepackage{booktabs}

\newcommand{\modelname}{\text{P4P}}
\newcommand{\modelnamespace}{\text{P4P} }

\title{\LARGE \bf
P4P: Conflict-Aware Motion Prediction for Planning\\ in Autonomous Driving
}

\author{Qiao Sun$^{1}$, Xin Huang$^{2}$, Brian C. Williams$^{2}$, and Hang Zhao$^{1}$ 
 \thanks{$^{1}$ Qiao Sun (\texttt{alan.qiao.sun@gmail.com}) and Hang Zhao (\texttt{hangzhao@mail.tsinghua.edu.cn}) are affiliated with IIIS, Tsinghua University and Shanghai Qi Zhi Institute.
 }
\thanks{$^{2}$ Xin Huang and Brian C. Williams are affiliated with CSAIL, Massachusetts Institute of Technology.}
}

\begin{document}


\maketitle
 
\begin{abstract}
Motion prediction is crucial in enabling safe motion planning for autonomous vehicles in interactive scenarios. It allows the planner to identify potential conflicts with other traffic agents and generate safe plans. Existing motion predictors often focus on reducing prediction errors, yet it remains an open question on how well they help identify the conflicts for the planner. In this paper, we evaluate state-of-the-art predictors through novel conflict-related metrics, such as the success rate of identifying conflicts. Surprisingly, the predictors suffer from a low success rate and thus lead to a large percentage of collisions when we test the prediction-planning system in an interactive simulator. To fill the gap, we propose a simple but effective alternative that combines a physics-based trajectory generator and a learning-based relation predictor to identify conflicts and infer conflict relations. We demonstrate that our predictor, \modelname, achieves superior performance over existing learning-based predictors in realistic interactive driving scenarios from Waymo Open Motion Dataset.

\end{abstract}

\section{Introduction}



Motion prediction plays an important role in enabling safe and efficient planning for autonomous vehicles. The prediction supports the planner to identify potential conflicts with other traffic agents on the road and make safe decisions to avoid collisions. The conflicts are commonly defined in transportation systems if two agents approach each other that could lead to a collision if their movements stay unchanged~\cite{perkins1968traffic,sayed1999traffic}. In a typical intersection example, the ego self-driving car needs to infer all possible conflicting agents, by predicting their future trajectories, in order to decide whether to keep moving or yield if needed. 

Recently, learning-based trajectory predictors have been studied extensively to reduce prediction errors and improve coverage~\cite{cui2019multimodal,gao2020vectornet,kim2021lapred,gu2021densetnt,sun2022m2i}, yet it remains unclear how well the predictions could support the planner to find potential conflicts with other traffic agents to generate safe and efficient plans~\cite{ivanovic2021injecting,huang2021tip}. In this paper, we study this key question by evaluating a prediction-planning system, which includes a planner that generates plans based on the prediction results from a predictor, in a closed-loop test bed. We gauge the performance of the predictor using novel conflict-related metrics, such as the success rate of identifying the conflicts. Our test bed leverages a state-of-the-art simulator~\cite{sun2022intersim} that simulates realistic and reactive agent behaviors in interactive scenes, which allows us to faithfully evaluate and compare the performance of prediction-planning systems that use various kinds of predictor models.

\begin{figure}[t]
    \centering
    \includegraphics[width=0.5\textwidth]{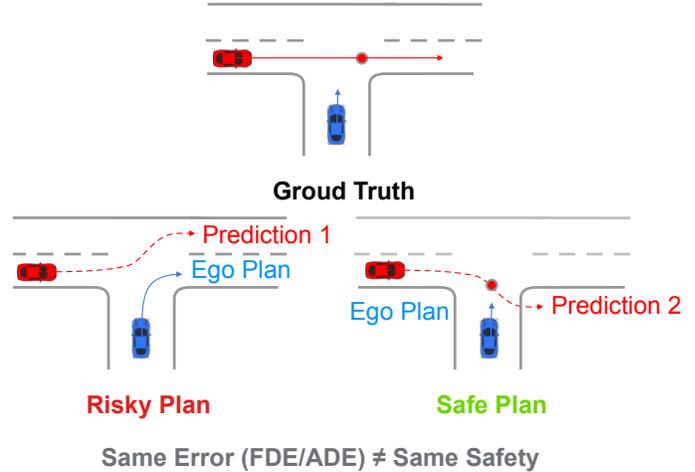}
    \caption{An example of two symmetric trajectory predictions with the same prediction errors having different effects on the downstream planner. Upper: The turning ego vehicle in blue is yielding to the red agent. Lower left: Prediction 1 fails to identify a critical conflict and leads to a dangerous plan for the blue ego vehicle. Lower right: Prediction 2 identifies the conflict that leads to a safe plan for the ego vehicle.}
    \label{fig:motivating_example}
\vspace{-6mm}
\end{figure}

The study shows that, surprisingly, the planner suffers from a high-collision rate in interactive scenarios given the predictions produced by state-of-the-art learning-based predictors. Due to the input noises and out-of-distribution inputs, these predictors tend to produce false or unstable predictions that miss the critical conflicting trajectories of traffic agents and thus lead to unsafe colliding ego plans. We present a toy example in Fig.~\ref{fig:motivating_example}, in which the false predictions for the red agent lead to an unsafe ego plan. 

In order to improve the robustness of the predictions to better serve the planner, we follow~\cite{scholler2020constant} and propose a simple but effective physics-based predictor that predicts consistent and reliable future trajectories of traffic agents and generalizes well to unseen scenarios. In the experiments, we show that despite having larger average prediction errors, the physics-based predictor recognizes more conflicting trajectories than the state-of-the-art learning-based predictors.

One caveat of the physics-based predictor is that it does not account for interactions among agents. For instance, a traffic agent should yield to another agent who has right of the way when a conflict happens. Therefore, we leverage a learning-based relation predictor to infer the pass/yield relation between the traffic agent and the ego vehicle when there exists a conflict. When the agent is predicted to yield, we refine its prediction from the physics-based model to slow down, which allows the planner to generate efficient ego plans to pass. On the other hand, we keep the conflicting prediction that would allow the ego vehicle to generate safe plans to avoid collisions, as illustrated in Fig.~\ref{fig:motivating_example}. 

\textbf{Contributions:} Our contributions are three-fold. First, we explore a key question in trajectory prediction, by studying how well the existing learning-based predictors could help the planner to identify conflicts with other agents, through a closed-loop test bed based on a state-of-the-art interactive simulator. We propose novel metrics to quantify the success rate of identifying conflicts and inferring conflict relations. Second, we investigate the common failure cases of the prediction-planning system and demonstrate that the missing conflicts inevitably lead to unsafe colliding plans. Third, we propose a simple but effective predictor that combines a physics-based trajectory generator and a learning-based relation predictor to overcome the gaps in existing predictors. We show that our approach achieves better results in supporting the planner to produce safer and more efficient plans in interactive driving scenarios.

\section{Related Work}
\vspace{-1mm}
\label{sec:related_work}
\subsection{Prediction for Planning}
\label{sec:work_pp}

Recently, learning-based motion predictors are proposed to predict future agent behaviors in complex scenarios~\cite{cui2019multimodal,gao2020vectornet,kim2021lapred,gu2021densetnt,sun2022m2i}. Despite their success in challenging prediction benchmarks, most predictors are evaluated by prediction accuracy and their contributions to the downstream task such as planning remain an open question~\cite{ivanovic2021injecting,huang2021tip}.

To answer this question, a few work has been proposed to evaluate the predictor in terms of the planner performance. For instance,~\cite{ivanovic2020mats,chen2022scept} examine their predictors in a combined prediction-planning framework, in which a model predictive controller leverages the predicted trajectory to plan socially-aware trajectories. Qualitative examples are presented to demonstrate the effectiveness of the proposed predictors. Recently,~\cite{huang2021tip} investigates the performance of different predictors by measuring quantitative decision making performance, in terms of the success rate of finding the optimal decision and the false alarm rate given the predictions. As these studies present insightful results on the performance of existing learning-based predictors in terms of supporting downstream planners, they often assume that the behavior of the surrounding agent is independent of the ego plan, by replaying the logged trajectories from data. In practice, the future trajectories of the surrounding agents change over time as the agents react to the ego plan, making such open-loop evaluations questionable. In this work, to faithfully evaluate a prediction-planning system, we leverage a closed-loop test bed based on a state-of-the-art simulator that simulates realistic behaviors of traffic agents in interactive scenarios~\cite{sun2022intersim}. 

\subsection{Closed-Loop Prediction-Planning Evaluation}
Evaluating the performance of prediction-planning systems in real-world driving scenarios is a non-trivial task, as it requires a tremendous amount of time and resources to build a testing fleet. More importantly, real-world tests could lead to accidents in interactive scenarios if the planner becomes too aggressive or misunderstands the future intention of other agents. This motivates the need to evaluate the planner in a simulated test bed, which is more cost-effective and covers a diverse set of scenarios, including risky ones. 

Existing closed-loop test beds often leverage simplified rule-based models, such as the intelligent driver model, to simulate the behaviors of nearby traffic agents~\cite{dosovitskiy2017carla,lopez2018microscopic,bernhard2020bark,caesar2021nuplan}, making it hard to evaluate planner performance in highly interactive scenes. Recently, various learning-based models~\cite{bergamini2021simnet,suo2021trafficsim,igl2022symphony,sun2022intersim} are proposed to simulate realistic and reactive agent behaviors to better support planner evaluation. In this work, we leverage a test bed that includes diverse and challenging driving scenarios with realistic agent reactions powered by a state-of-the-art interactive simulator~\cite{sun2022intersim}. The test bed allows us to compare planning-prediction systems that use different prediction models and find the failure cases, which are rarely discussed in existing learning-based prediction work.

\subsection{Interaction Relation Prediction}
Predicting interaction relations has demonstrated to be effective in improving performance of interactive trajectory prediction~\cite{kumar2020interaction,ban2022deep,sun2022m2i}. The predicted relations are useful in generating consistent trajectories among multiple agents in interactive scenarios. In this paper, we leverage this key insight and predict the interaction relations between the ego vehicle and environment agents to support safe and efficient planning when a conflict exists. By combining the relation predictor with a physics-based trajectory generator, our approach has demonstrated to outperform complicated learning-based trajectory predictors in terms of identifying conflicts and supporting the planner to generate safe plans.

\section{Problem Formulation}
\label{sec:formulation}

In this section, we introduce the problem formulation for our prediction-planning system, including a conflict-aware trajectory predictor that predicts future trajectories for other traffic agents and a planner that generates safe ego plans given the predictions.

\subsection{Conflict-Aware Trajectory Prediction}
In this work, we follow~\cite{huang2021tip} and define the following trajectory prediction problem. Given observed states $X$, including the map states and the observed agent states, and an ego plan $P$, the goal is to predict a set of $K$ future trajectory samples $Y$ up to a finite horizon $T$ for the agents. Compared to existing prediction models, our predictor accounts for the potential conflicts with the ego plan. We generalize the definition of conflict at intersections in~\cite{perkins1968traffic,sayed1999traffic} and define a conflict to exist if the trajectories of the ego plan and the traffic agent cross \textit{in space}, as customary in recent interaction prediction and simulation literature~\cite{kumar2020interaction,ban2022deep,sun2022m2i,sun2022intersim}. Note that a conflict does not necessarily imply a collision, which happens if two trajectories cross in \textit{both time and space}. 
In practice, it is often important to identify conflicts with nearby traffic agents for the ego vehicle as these conflicts could lead to a collision if the planner ignores the conflict.

\subsection{Planning}
In the planning task, we assume that a pre-defined trajectory is given to the planner. The pre-defined trajectory is generated without considering the collisions with other traffic agents. Such a trajectory can be obtained efficiently from an interaction-ignorant planner that only considers static obstacles~\cite{gonzalez2015review,fan2018baidu}. Given the pre-defined trajectory, the planner chooses from two actions: following the pre-defined trajectory, and slowing down along the path of the pre-defined trajectory to yield to traffic agents whose predicted future trajectories indicate a collision.

The optimal action is determined by the simulated future trajectories of nearby traffic agents. If they cause a collision with the pre-defined ego trajectory, the optimal action is to slow down and vice versa. The objective of the planner is to choose the optimal action given the predictions.

\section{Methods}
\label{sec:methods}


\begin{figure}[t]
    \centering
    \includegraphics[width=0.4\textwidth]{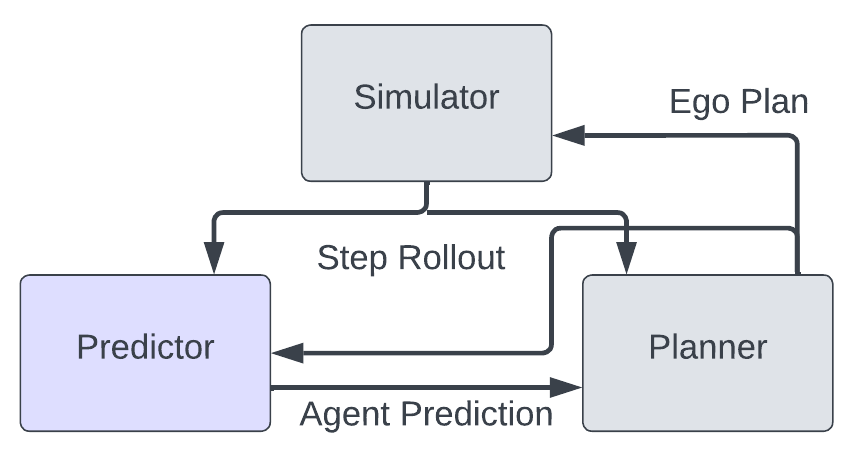}
    \caption{Overview of \modelnamespace Test Bed, which includes a predictor that predicts conflict-aware future trajectories of traffic agents, a planner that generates ego plans given the predictions, and a simulator that rolls out agent trajectories based on the ego plan and the simulated agent trajectories.}
    \label{fig:test_bed}
\vspace{-4mm}
\end{figure}

Our proposed method includes two core components.
In the first component, we introduce a closed-loop test bed through an interactive simulator to evaluate motion predictors in a prediction-planning system. The performance is measured by a few novel conflict-based metrics.
In the second component, we propose our conflict-aware trajectory predictor that combines a physics-based trajectory generator and a learning-based relation predictor.
In the remaining of this section, we present each component in more detail.


\subsection{Test Bed for Prediction-Planning Systems}
\label{sec:pred_eval}

\subsubsection{Test Bed Overview}
As shown in Fig.~\ref{fig:test_bed}, the test bed includes a \textit{predictor} that predicts conflict-aware future trajectories of traffic agents, a \textit{planner} that generates safe plans given the predictions, and a \textit{simulator} that simulates realistic and reactive agent behaviors over time. We run the planner in the test bed over a fixed horizon and identify collisions between the ego plan and the simulated behaviors of traffic agents for quantitative and qualitative safety analysis\footnote{We will open-source our closed-loop test bed that includes a diverse set of realistic and interactive scenarios and encourage the community to benchmark their prediction-planning systems on it. While we fix the planner in this work and focus on evaluating the system performance with different predictors, one can use our test bed to evaluate various planners.}.

\subsubsection{Planner}
\label{sec:planner}
We follow existing motion planning literature~\cite{gonzalez2015review,fan2018baidu} and generate future ego trajectories based on its possible routes given the map information\footnote{While there exists a wide range of planners in the field, this paper focuses on evaluating the performance of the trajectory predictor and thus uses a minimal planner for the sake of simplicity. We defer to more advanced planners for future work.}. A route is defined as a series of lanes to follow. To find a route of the ego vehicle, we first identify the closest lane to itself based on Euclidean distance. Next, we search over the connecting lanes iteratively to obtain a set of possible routes to follow. We assume the goal to be given for the ego vehicle and choose the route that passes the goal point with the shortest distance. If there is no such route, we randomly select one route from the set to follow. 

To obtain a continuous trajectory from a route, we use Cubic Bézier curves~\cite{liang2012automatic, gonzalez2014continuous} to connect the starting location of the agent and the starting lane and to connect consecutive lanes. The speed of the trajectory is set to the speed limit of the lane. If the starting speed is different from the speed limit, we change the speed at an acceleration of $0.3m/s^2$ or a deceleration of $-0.75m/s^2$ until it reaches the speed limit. Despite the simplicity of our route-based planner, we find it to be effective in generating a smooth and efficient path to the goal point, as shown in the experiments.

To handle collision avoidance, we further revise the plan given the predictions of the traffic agents. If any predictions indicate a collision with the ego plan, we revise the speed profile over the planned trajectory to slow down before the earliest cross point. If the cross point is less than 2 meters away, we apply an emergency stop with a deceleration of $-1.5m/s^2$. Otherwise, we apply a smaller deceleration of $-0.75m/s^2$ to balance driving efficiency and comfort. On the other hand, if no collisions are indicated by the predictions, we simply execute the original ego plan.

\begin{figure*}[t]
    \centering
    \includegraphics[width=1.0\textwidth]{images/fig_copred.pdf}
    \vspace{-6mm}
    \caption{Overview of \modelnamespace Predictor. It first generates possible future trajectory candidates of traffic agents through a physics-based model and routes information when available. If a conflict is detected with the ego plan, we leverage a relation predictor to infer whether the conflicting agents are yielding and refine the predictions to resolve the conflict if necessary.}
    \label{fig:CONTACT}
\vspace{-6mm}
\end{figure*}

\subsubsection{Interactive Simulator}
We propose a closed-loop test bed to evaluate the performance of a prediction-planning system through a large number of testing scenarios. To encourage coverage in our test bed, we initialize each scenario by a data sample in the Waymo Open Motion Dataset (WOMD)~\cite{ettinger2021large}, a large-scale dataset collected from real-world urban driving. A key challenge in the test bed is to simulate realistic traffic agent behaviors that react to the ego vehicle. Existing works often rely on log replay to simulate traffic agent trajectories based on the observations from data or use simple heuristic-based agent models. In this work, we use a state-of-the-art simulator~\cite{sun2022intersim} to simulate agent behaviors with realistic reactions based on a learned agent model. It supports diverse reactions by randomly selecting a feasible route for each traffic agent and generating diverse curvatures for turning trajectories by randomizing the connecting points between consecutive lanes.

\subsubsection{Evaluation Metrics}
\label{sec:metrics}
To quantify how well the predictor supports the planner in terms of identifying conflicts and inferring conflict relations, we propose a few metrics to evaluate the performance of the prediction-planning system:
\begin{enumerate}
    \item \textbf{Conflict Prediction Recall}: The number of successfully identified conflicts given the predictions over the total number of conflicts happened in the test bed. This measures the ability of the predictor to support the planner in terms of identifying conflicts.  A conflict exists if the ground truth future trajectories of traffic agents cross with the ego plan in space based on the bounding boxes of the agents. Examples of conflicts are illustrated as red dots in Fig.~\ref{fig:CONTACT}.
    \item \textbf{Conflict Relation Accuracy}: The percentage of correctly identified relations \textit{given a conflict is successfully identified}. The relation includes yield or pass, based on which agent arrives at the cross point first~\cite{sun2022m2i}. In safety critical scenarios, a planner not only has to identify potential conflicts for safe driving, but also needs to know whether the other agents are yielding to maintain efficiency.
\end{enumerate}

In addition to the conflict-aware metrics, we use the following metrics to measure safety and efficiency based on existing literature~\cite{suo2021trafficsim,bergamini2021simnet}.
\begin{enumerate}
    \setcounter{enumi}{2}
    \item \textbf{Collision Rate:} The number of collisions involving the ego vehicle divided by the total number of scenarios. Compared to existing work~\cite{ivanovic2020mats,chen2022scept,huang2021tip}, we measure this metric in a closed-loop test bed.
    \item \textbf{Progress:} The total meters traveled by the ego vehicle without being terminated divided by the total number of scenarios. We terminate the simulation if a collision happens between the ego vehicle and a traffic agent.
    \item \textbf{Stuck Rate:} The ratio of scenarios in which the ego vehicle does not move over the entire planning horizon.  
\end{enumerate}


\subsection{Conflict-Aware Prediction}
\label{sec:pred_eval}

\subsubsection{Predictor Overview}
The objective of our predictor is to generate stable and consistent future trajectories of traffic agents that help the planner identify a conflict if it exists. As shown in Fig.~\ref{fig:CONTACT}, we combine a physics-based trajectory generator to provide stable trajectories with sufficient coverage, and a learning-based relation predictor to help refine the prediction in interactive scenarios.

\subsubsection{Physics-Based Trajectory Generation}
\label{sec:traj_gen}
We follow~\cite{scholler2020constant} and adopt a physics-based approach to generate possible future trajectories for the traffic agents, by assuming the agent travels at a constant velocity along its path. For agents such as vehicles and (motor)cyclists, we can obtain their possible paths given the map information. For agents such as pedestrians, we assume that the agent maintains its heading direction with small perturbance to provide multiple paths. Compared to a complicated learning-based motion predictor, we find our simple approach produces consistent and stable future trajectories that cover multiple agent intent modalities and help the planner to identify conflicts when they exist.

\subsubsection{Relation Prediction}
As the physics-based trajectory generator does not account for interactions with the ego vehicle, it can produce false predictions in which the traffic agent is predicted to cause a conflict with the ego plan, but in reality the traffic agent is yielding. We can typically identify these false predictions through trivial heuristics based on traffic rules, yet there are many nontrivial cases in which there are no explicit traffic rules or driving norms. Therefore, we propose to use a learning-based relation predictor to classify the relation between the ego vehicle and the traffic agents in ambiguous cases. We adopt the relation prediction model from a state-of-the-art multi-agent predictor~\cite{sun2022m2i}. The predictor leverages a backbone based on VectorNet~\cite{gao2020vectornet} that encodes context information, such as observed trajectories and map data, through a vectorized representation and uses an MLP layer to predict the interacting relation as a binary yield/pass classification problem. 

Compared to a learning-based trajectory predictor, our relation predictor has the following advantages. First, it affords a smaller prediction space, as it only predicts the relations as a binary classification problem as opposed to a complicated continuous regression problem for trajectory prediction. A smaller prediction space essentially reduces the complexity of the problem, making it less sensitive to noisy inputs and taking less inference time. Second, our relation predictor offers better interpretability -- its results are easier to interpret than complicated trajectories. Third, as our predictor outputs a distribution over relations, we can set different thresholds to tune the level of conservativeness of ego behaviors when yielding to other traffic agents. 

\subsubsection{Prediction Refinement}
In cases where the traffic agent is predicted to yield, we refine its prediction to slow down and stop before the cross point with the ego plan, thus resolving the conflict with the ego vehicle. Otherwise, we keep the original predictions from the physics-based trajectory generator and defer to the ego planner to resolve the conflict.

\begin{table}[t!]
    \centering
    \footnotesize
    \bgroup
    \begin{tabular}{lcccc}
    \toprule
    \makecell{WOMD \\ Interactive \\ Dataset} & \makecell{No \\ Conflicts} & \makecell{Trivial \\ Conflicts \\ (Blocking)} & \makecell{Trivial \\ Conflicts \\ (Traffic Lights)} & \makecell{Non-Trivial \\ Conflicts} \\
    \midrule
    \makecell{Training}& 13.46\% & 48.54\% & 27.27\% & 10.73\% \\
    \makecell{Validation}& 3.12\% & 53.30\% & 29.30\% & 14.28\% \\
    \bottomrule
    \end{tabular}%
    \egroup
    \caption{Summary of conflicts in the interactive WOMD.}
    \label{tab:dataset}
\vspace{-6mm}
\end{table}

\section{Experiments}
\label{experiments}
We use the Waymo Open Motion Dataset (WOMD)~\cite{ettinger2021large} to initialize the scenarios in our test bed and to train and validate our relation predictor. The interactive WOMD includes 204,166 scenarios in the training set and 43,479 scenarios in the validation set. To verify the density of interactions, we measure the percentage of conflicts between the logged trajectories in the dataset, based on the definition in Sec.~\ref{sec:formulation}. We further divide the conflicts into trivial conflicts, which can be identified by traffic rules or blocking relations, i.e. one agent blocking the path of the other, and non-trivial conflicts that require a more complicated learning-based method to identify. The statistics are presented in Table~\ref{tab:dataset}. 
In the experiments, we use the scenarios in the validation set that exhibit non-trivial conflicts to evaluate the performance of the planning-prediction system.

\subsection{Test Bed Setup}
We initialize our test bed based on the WOMD validation set with nontrivial conflicts. In each scenario, we randomly select one agent as the ego vehicle from the agents of interest marked by the dataset.

At each time step during evaluation, the predictor predicts 6 trajectory samples over the next 8 seconds based on 1.1 seconds of observations. The planner then generates the ego plan over the next 8 seconds given predictions. The prediction and planning frequency is 10Hz. The ego plan is used to roll out the position of the ego vehicle at the next time step. To roll out the trajectory of the traffic agents, we use a state-of-the-art interactive simulator~\cite{sun2022intersim} that simulates reactive agent behaviors.

\subsection{Learning-Based Relation Predictor}
We implement our relation prediction following~\cite{sun2022m2i}, except that we modify the relation predictor head to output the probability over two possible outcomes -- yield and pass, based on which agent passes the cross point first.
We train our relation predictor model using 8 Nvidia 3080 GPUs with a batch size of 64 for 60 epochs. The accuracy on the interactive training set is $95.65\%$, and the accuracy on the interactive validation set is $89.29\%$ with an F1 score of $0.58$. 

\subsection{Baselines}
We present a few representative predictors to compare the performance with our proposed method.
    \textbf{Constant Velocity}~\cite{scholler2020constant}: A physics-based predictor that predicts future trajectories based on a constant velocity model. Despite its simplicity, it produces stable results given noisy inputs and generalizes well to unseen scenarios.
    \textbf{TNT}~\cite{zhao2020tnt}: An anchor-based trajectory predictor model that has achieved top performance in various trajectory prediction benchmarks such as Argoverse~\cite{chang2019argoverse}, INTERACTION~\cite{zhan2019interaction}, and SDD~\cite{robicquet2016forecasting}.
    \textbf{DenseTNT}~\cite{gu2021densetnt}: A state-of-the-art trajectory predictor model on WOMD based on an anchor-free approach.
In addition to existing predictors, we propose a variant of our model, \textbf{\modelname-NoRelation} that does not use the relation predictor to refine the predictions if the traffic agents are predicted to yield. This predictor would result in conservative plans that always yield regardless of the conflict relation.

\begin{figure*}[t]
    \centering
    \includegraphics[width=0.95\textwidth]{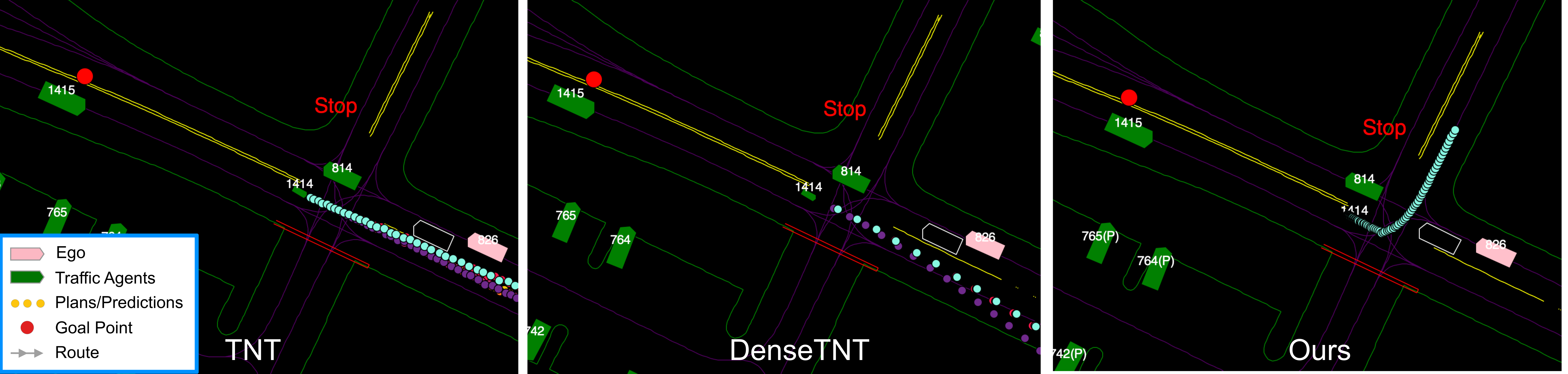}
    \caption{Left and Middle: TNT and DenseTNT generate false predictions that miss a critical conflict between the ego vehicle and the left turning cyclist. Right: \modelnamespace generates trajectories that successfully identify the conflict, resulting in a safe ego plan that yields to the cyclist.}
    \label{fig:compare} 
\end{figure*}

\begin{figure*}[t]
    \centering
    \includegraphics[width=0.95\textwidth]{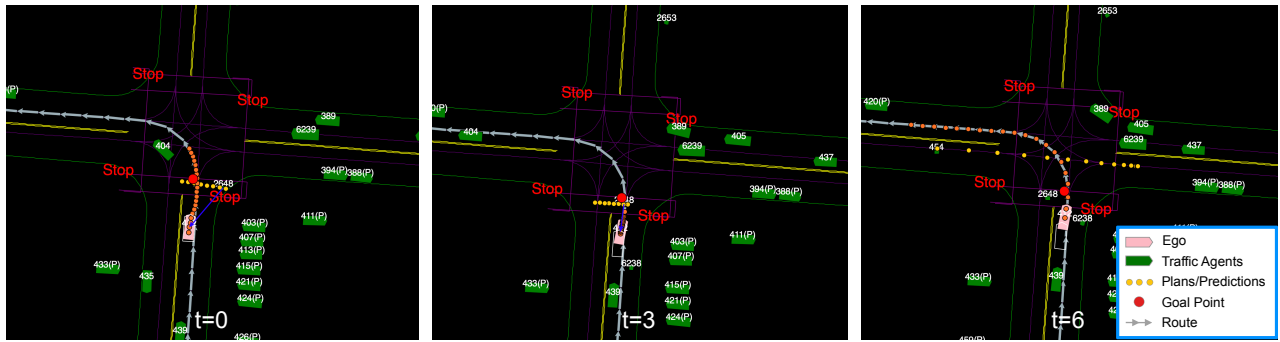}
    \caption{\modelnamespace successfully identifies conflicts with multiple agents over time, resulting in safe plans at 3s and 6s.}
    \label{fig:pred1}
\end{figure*}

\begin{table*}[t!]
    \centering
    \footnotesize
    \bgroup
    \begin{tabular}{lccccc}
    \toprule
    
    Methods & \makecell{minFDE[m]} $\downarrow$ & \makecell{minADE[m]} $\downarrow$ & \makecell{Conflict \\ Prediction Recall \\ (Top 1)} $\uparrow$ & \makecell{Conflict \\ Prediction Recall \\ (Top 6)} $\uparrow$ & \makecell{Conflict Relation\\ Prediction Accuracy} $\uparrow$  \\
    
    \midrule
    
    Constant Velocity
    & 17.29 & 6.53 & 44.08\% & 75.40\% & 69.01\% \\
    TNT & 16.02 & 4.70 & 16.13\% & 23.49\% & 38.46\% \\
    DenseTNT & \textbf{3.65} & \textbf{1.74} & 14.01\% & 22.11\% & 57.32\% \\
    \modelname-NoRelation & 16.85 & 6.22 & 57.73\% & 89.34\% & 69.52\% \\
    \modelnamespace & 15.67 & 5.87 & \textbf{58.20\%} & \textbf{97.07\%} & \textbf{71.78\%} \\
    

    
    \bottomrule
    \end{tabular}
    \egroup
    \caption{Predictor performance comparison. Compared to learning-based predictors, physics-based predictors are better at identifying conflicts and inferring conflict relations, despite having higher prediction errors. Our model outperforms a simple constant velocity model by leveraging a relation predictor and additional route information.}
    \label{tab:result}
\end{table*}

\subsection{Quantitative Prediction Performance Analysis}
In Table~\ref{tab:result}, we present a quantitative comparison between our proposed predictor and the baseline predictors in terms of prediction accuracy and the ability to identify conflicts and conflict relations. The results show that despite having lower prediction errors, such as minADE and minFDE~\cite{gupta2018social}, the learning-based predictors such as TNT and DenseTNT suffer from poor conflicts prediction recall and relation prediction accuracy. On the other hand, a simple physics-based predictor such as Constant Velocity achieves much better performance in identifying conflicts and inferring conflict relations. Our proposed method further improves the performance by leveraging a relation predictor and route information, and achieves the best performance among baseline predictors. 

\subsection{Quantitative Planning Performance Analysis}
In Table \ref{tab:simulation}, We present a quantitative analysis of the planner that uses our predictor and the baseline predictors. 
Results show that the planner that uses physics-based models achieves better safety (lower collision rate) and efficiency (higher progress) compared to learning-based predictors. Compared to physics-based baselines, our proposed method achieves better efficiency in terms of progress and stuck rate by leveraging a relation predictor to generate pass plans for the ego vehicle when the traffic agents are predicted to yield in the presence of conflicts.

\begin{table}[t!]
    \centering
    \footnotesize
    \bgroup
    \begin{tabular}{lccc}
    \toprule
    Predictors & \makecell{Collision Rate} $\downarrow$ & \makecell{Progress[m]} $\uparrow$ & \makecell{Stuck Rate} $\downarrow$ \\
    \midrule
    Constant Velocity & \textbf{1.29\%} & 24.50 & 3.55\% \\
    TNT & 14.10\% & 24.58 & \textbf{2.08}\% \\
    DenseTNT & 12.49\% & 21.59 & 4.00\% \\
    \modelname-NoRelation & 1.59\% & 24.57 & 3.34\% \\
    \modelnamespace & 1.39\% & \textbf{24.84} & 3.07\% \\
    \bottomrule
    \end{tabular}%
    \egroup
    \caption{Closed-loop simulation results of the prediction-planning system using different predictors. Our proposed model achieves much better safety performance compared to learning-based predictors while having comparable efficiency performance in terms of progress and stuck rate.}
    \label{tab:simulation}
\end{table}

\subsection{Qualitative Analysis}

We introduce two representative qualitative examples\footnote{More examples can be found in the supplementary material.} to showcase the advantage of \modelnamespace in predicting conflict-aware trajectories and supporting the planner.

\textbf{Predict Conflict-Aware Trajectories}: In the example in Fig.~\ref{fig:compare}, the learning-based TNT and DenseTNT predictors fail to cover a safety-critical conflict for the left turning cyclist (agent 1414 in green) whose observed trajectory is noisy, leading to a colliding plan for the ego vehicle (in pink). More specifically, we observe that the DenseTNT generates unrealistic prediction samples (in colored dots) given the noisy inputs. On the other hand, \modelnamespace generates prediction samples that successfully identify the conflict and result in a safe ego plan.


\textbf{Support Safe Planning}: In an interactive intersection scenario in Fig.~\ref{fig:pred1}, our proposed method \modelnamespace successfully identifies the conflicts with the coming pedestrian and helps the planner to generate a yielding plan at 3s. It also identifies the conflict with the coming motorcycle, making the planner slow down again at 6s as the ego vehicle continues driving forward, after the conflict with the pedestrian is clear.

\section{Conclusions}
\label{conclusions}
In summary, we use a closed-loop test bed based on an interactive simulator to examine the capability of state-of-the-art learning-based predictors in terms of supporting the ego vehicle planner. Surprisingly, the planner suffers from a high collision rate due to the false predictions that ignore the conflict with the ego plan. We propose a simple but effective alternative to generate stable predictions that are effective at identifying conflicts and inferring conflict relations. Our approach \modelnamespace combines a physics-based trajectory generator and a learning-based relation predictor. Experiments in realistic interactive scenarios show that \modelnamespace achieves better safety and efficiency compared to the baselines.

\bibliographystyle{IEEEtran}
\bibliography{references} 

\end{document}